# Transfer Learning for Real-time Deployment of a Screening Tool for Depression Detection Using Actigraphy

Rajanikant Ghate, Nayan Kalnad
Avegen Ltd, London, UK
rajanikant@avegenhealth.com,
nayan@avegenhealth.com

Rahee Walambe, Ketan Kotecha
Symbiosis Institute of Technology
Symbiosis Centre for Applied Artificial Intelligence,
Symbiosis International University, Pune, India
rahee.walambe@sitpune.edu.in,
director@sitpune.edu.in

*Abstract*— Automated depression screening and diagnosis is a highly relevant problem today. There are a number of limitations of the traditional depression detection methods, namely, high dependence on clinicians and biased self-reporting. In recent years, research has suggested strong potential in machine learning (ML) based methods that make use of the user's passive data collected via wearable devices. However, ML is data-hungry. Especially in the healthcare domain primary data collection is challenging. In this work, we present an approach based on transfer learning, from a model trained on a secondary dataset, for the real-time deployment of the depression screening tool based on the actigraphy data of users. This approach enables machine learning modelling even with limited primary data samples. A modified version of leave-one-out cross-validation approach performed on the primary set resulted in mean accuracy of 0.96, where in each iteration one subject's data from the primary set was set aside for testing.

## I.  INTRODUCTION

Despite recent years witnessing a significant increase in depression screening, there is still scope for improvement [1,2]. The importance of early detection of depression in achieving better treatment outcomes [3,4] as well as WHO's report about the prevalence and consequences of depression [5] also support the need for more screening [6].

The bottlenecks in depression screening can be attributed to the presence of self-reporting bias as well as the extensive requirement of clinical experts for contemporary screening and diagnosis methods [7,8]. Such screening relies heavily on the ability, desire and honesty of a patient.
This suggests a need for a screening tool that is:
- Unbiased to user reports
- Have lesser dependence on clinical expertise
- Doesn't require users' active inputs

Since 2015, an emerging science called 'digital phenotyping' has demonstrated that patterns in interactions of human beings with digital devices can be studied for their health status identification [9]. Of the various possible modalities and streams of passive data, the most interesting is actigraphy data [10]. Actigraphy data is typically total or continuous tracking of activity or movement, most commonly tracked by a smartwatch attached to a wrist. It is non-invasive, continuously tracked, doesn't intrude on privacy and isn't dependent on user's reporting bias. There are strong pieces of evidence to suggest that beyond the established use of actigraphy to monitor intricate circadian rhythms and energy fluctuations [11] actigraphy can also have strong potential to differentiate a healthy person from one living with depression [12,13].

Machine Learning (ML) has gained popularity for its ability to discover complex patterns in data [14]. In this context of depression detection, research suggests that passive sensing data can be coupled with advanced machine learning techniques [14-17].

However, ML is data-hungry [18]. For models to be generalisable, and to be clinically reliable, the dataset must be substantially large. Let aside the heavy requirements for model training, even for the right clinical validation data size sufficiency criteria need to be met by statistical considerations.

The required data can be collected from scratch through an independent research project, or an existing dataset can be used, of course, with due respect to the terms and conditions of usage. While the data collected through the first approach has been referred to as "Primary" in our work, the data that is available in the open domain has been referred to as "Secondary".

Each method of data collection has its own merits and limitations. Researchers can have full control over specifications as well as the quality of a primary dataset. However, building a primary dataset in healthcare is not just expensive but also draws a need for ethical clearance as it involves human beings. Also, building a primary dataset takes a considerable amount of time depending on the study duration and ease of recruitment. While using a secondary dataset can speed up the activity, a major challenge remains lack of control over "what" data is being collected. Often a single secondary dataset source might not be sufficient. In such a scenario another secondary dataset that is compatible with the previous one can be challenging to find. Also, since secondary dataset is not owned, due diligence on its usage terms and conditions must be carried out.

Considering our research scenario, in this paper, we have proposed a hybrid approach. We have first trained and cross-validated model on a secondary dataset named "Depresjon" [19]. Then with the help of a primary dataset consisting of very limited samples, transfer learning [20] has been applied to train serving models that use fitibit's steps count data as input for prediction. Transfer learning applies to an approach where model is trained on a source dataset, and is applied for predictions on a different target dataset.

A demonstration user interface (UI) has been set-up, that has these models deployed in the backend, allowing users to drop their fitbit steps counter data file for prediction.

In summary, the contribution is two-fold:
- A demonstration UI that illustrates the end-user, deployment methodology for depression detection models trained using passive data such as actigraphy.
- A novel yet simple approach of transfer learning, making collaborative research usable enabling working with a limited data samples as compared to traditional approach of extensive primary data collection. A collaborative approach will accelerate the setup of a generalized data.

## II. METHODS

### A. Data Collection – Primary and Secondary

For training, the secondary dataset called "Depresjon" [19] was downloaded. The data contained continous minute wise activity levels for 55 participants. The time longitude for each participant was varying, with an average of 12.6 days. Wrist-worn accelerometer sensor-based device called as Actiwatch has been used for the study. Of the 55 participants, 23 are labeled as living with depression while 32 are abeled as healthy. MADRS scale readings [21] is provided for all the participants.

For primary dataset collection, voluntary participation was sought. All the interested participants were explained the details and the importance of the study. The informed consent was obtained from all the final shortlisted participants. For primary dataset, fitbit step [22] counts logs have been collected from 4 participants. Each one of them were asked to self-evaluate their depression status using PHQ-9 questionnaire [23]. All these participants self-reported as healthy.

### B. Problem Construct and Target Variable

The problem statement can be defined in a number of different ways. Simplest being, every participant can be classified as a depressed or as healthy. However, in this approach much larger number participants data will be needed for any significant machine learning. [18].

Another practical possible approach is of classifying each separate block wise data of a particular participant. This block can be of 1 min, 1 day, 2 days and so on depending upon the granularity of data as well as longevity of data per participant. The model in this case then will predict if a certain piece of new data of similar length, would match to someone who is depressed or healthy.

Considering 24-hour data as minimum length required for defining circadian rhythm, in the proposed approach each participant's 24 hour data has been considered as a sample for classification problem. This implies that if a participant had 7 days of recording, then each of those 7 days would be treated as separate labels. Target variable assigned to these samples were '0' or '1' depending on whether the participant belonged to healthy or depressed group, respectively.

### C. Feature Engineering

Each participant's 24-hour data was aggregated to hourly total activity from minute wise activity. This aggregation was then binned into 4 sub-parts depending upon hour of the day. This resulted in 4 subsets of 6 elements each. Statistical features for each subset have been computed and finally collated as final features for ML experimentation. This approach has been illustrated in Fig. 1.

The rationale for such feature engineering is to observe the fluctuations in the activity levels at different times of the day. For example, if a person just woke up once during the sleep hours versus a person who was restless for significant periods would be very much evident from such features.

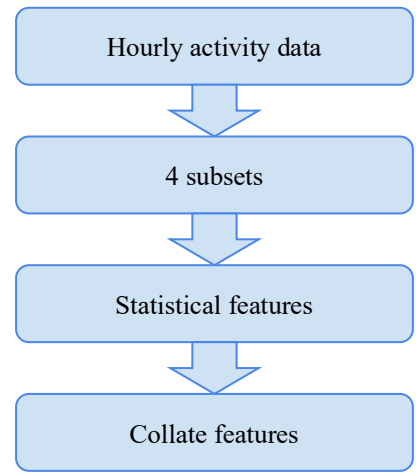

Figure 1. Feature engineering approach.

The hourly activity had missing values. On any day, if participant's missing values were for more than 2 hours, that day's data was dropped from creating features. However, all data has been used for scaling hourly activity. For the days where 22 or more hours were available, missing values have been imputed with average of previous and next available hourly activity.

### D. Modelling Approach

To begin with, a cross-validation experimentation of 5-fold was performed. Random forest model with class-weight hyperparameter as "balanced" was trained in each fold. In the same setup, dummy classifier with stratified strategy was implemented for baseline results estimate. Both the dummy and the random classifier has been built using scikit-learn [24]. This experiment was done to assess if the models trained from the secondary dataset have the ability to differentiate between the two groups.

Next, on the same dataset, a modified version of LOOCV was performed. In this setup, iteratively, data corresponding to a pair of participants (one healthy and one depressed) was left aside for testing, while the remaining data was used to train models. Such a setup would allow understanding whether the models are generalizable across people or are specific to people part of the training set.

The core challenge has been to develop an approach for training generalizable models despite differing actigraphy devices. In a standard data science approach, feature scaler or transformer is trained on train set and the same scaler is applied on the test set. However, this approach will not work here as the activity levels are not of same unit. For this we have suggested to normalize hourly activity levels separately for each device before creating features. Two of the commonly used scalers of sk-learn library "MinMax" and "Robust" scalers have been explored in this work.

In order to check the feasibility of generalizability we also studied Q-Q plots [Fig. 2] between hourly activities of secondary dataset and primary dataset across all individuals.

The validation of the proposed approach had been also done using modified LOOCV approach. In this setup, model has been trained on entire Depresjon dataset, after scaling of hourly activity levels. Then in each iteration of the setup, one subject's data from the primary set is used for testing. The hourly activity data for this participant has then scaled with respect to the primary set that has been set aside. Features for each day as suggested in feature engineering section have been computed using scaled hourly data. The results for each test participant's each day's data have been computed.

For evaluating performance, in each of the setup average of sensitivity, specificity and accuracy have been reported as sensitivity and specificity analysis is one of the most common performance measurement analysis for diagnostic as well as screening tools [25].

### E. Deployment

The model trained on scaled hourly data of Depresjon has been deployed on a stand-alone UI with the help of GRADIO [26]. GRADIO is a free-ware that allows easy construction of standard UI with python and deployment of ML models in the backend. Any user intending to use this app, will be able to upload a fitbit JSON file containing step count log. This file can be easily downloaded from weblogin of fitbit console [27]. Once the input file is submitted, the data is processed to the required format, and for the most recent 15 days where recording has more than 22 hours, depression status is displayed.

The entire deployment flow has been summarized in Fig. 3.

### III. RESULTS

Results for all ML modeling experiments are shown in Table 1. Models built on Depresjon data, have consistently beaten the baseline dummy classifier in both validation setups, that is 5-fold CV as well as modified LOOCV. The results suggests that there are clearly distinct patterns between healthy and non-healthy participants' hourly data observed on day level.

The average confusion matrix for 5-fold CV based on Depresjon's data has been reported in Table 2. Also, for the same setup, ROC curve of each fold as well as mean ROC curve has been reported in Fig 4.

The accuracy observed in LOOCV setup is lower than that of 5-fold CV setup. This is primarily because 4 out of 21 pairs in simulation had poor accuracy. This suggests that a small proportion of participants have patterns that are not generalizable from this data alone.

Data points on Q-Q plots between hourly activities of secondary dataset and primary dataset suggest strong correlation between percentiles. This suggests that despite both having significantly different unit scales, a mathematical transformation can bring both to same distribution. Of the two approaches for scaling, "Robust" scaler clearly stood out compared to "MinMax" scaler.

Table 1. Mean and standard deviation of metrics observed in ML experiments.

| Experiment | Sensitivity | | Specificity | | Accuracy | |
|---|---|---|---|---|---|---|
| | Mean | SD | Mean | SD | Mean | SD |
| **Jakobsen, P et al.** | 0.65 | NA | 0.78 | NA | 0.73 | NA |
| **Random classifier 5-fold CV** | 0.35 | 0.01 | 0.60 | 0.01 | 0.51 | 0.01 |
| **Depresjon 5-fold CV** | 0.59 | 0.12 | 0.89 | 0.01 | 0.79 | 0.05 |
| **Random classifier LOOCV** | 0.44 | 0.02 | 0.73 | 0.06 | 0.61 | 0.04 |
| **Depresjon LOOCV** | 0.51 | 0.35 | 0.86 | 0.22 | 0.71 | 0.18 |
| **Primary Data LOOCV** | NA | NA | 0.96 | 0.10 | 0.96 | 0.10 |

Table 2. Confusion matrix for Depresjon 5 fold CV

| | Predicted Positive | Predicted Negative |
|---|---|---|
| Actual Positive | 42 | 29 |
| Actual Negative | 14 | 121 |

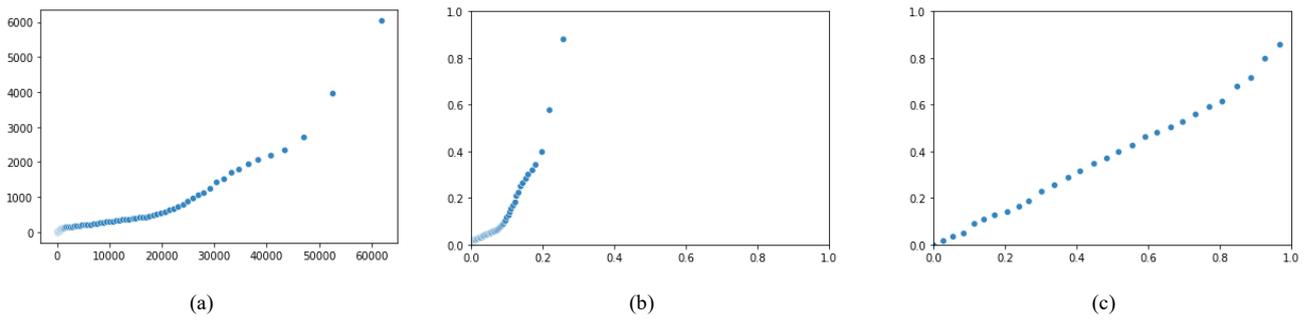

Figure 2. (a) Q-Q plot between hourly activity of Depresjon's participants (x-axis) and hourly step counts measured through fitbit (y-axis). (b) Q-Q plot after applying MinMax Scaler. (c) Q-Q plot after applying Robuts Scaler.

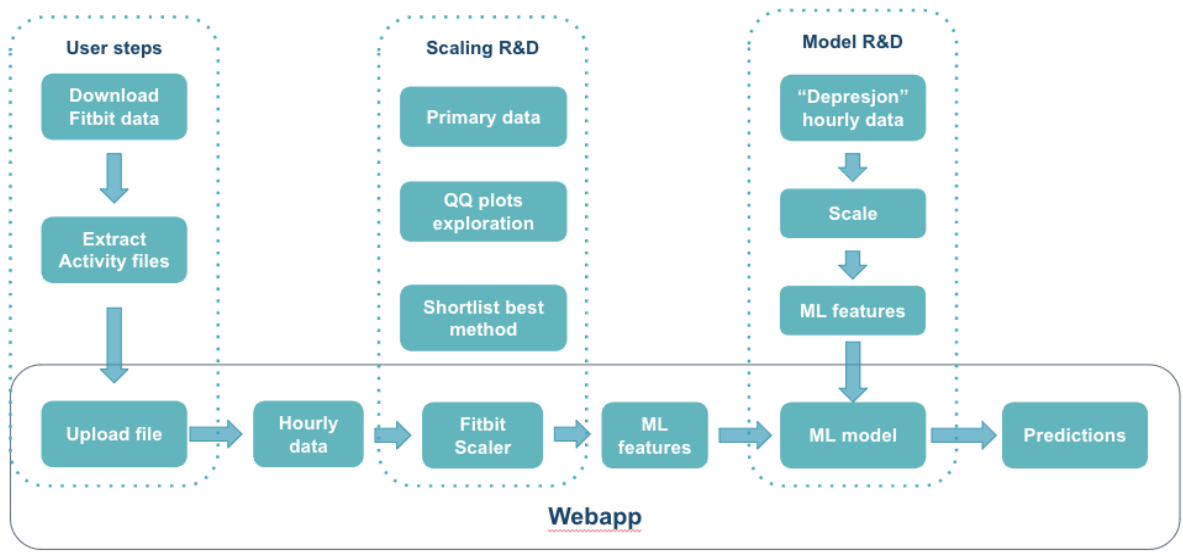

Figure 3. Research to deployment workflow of the proposed approach

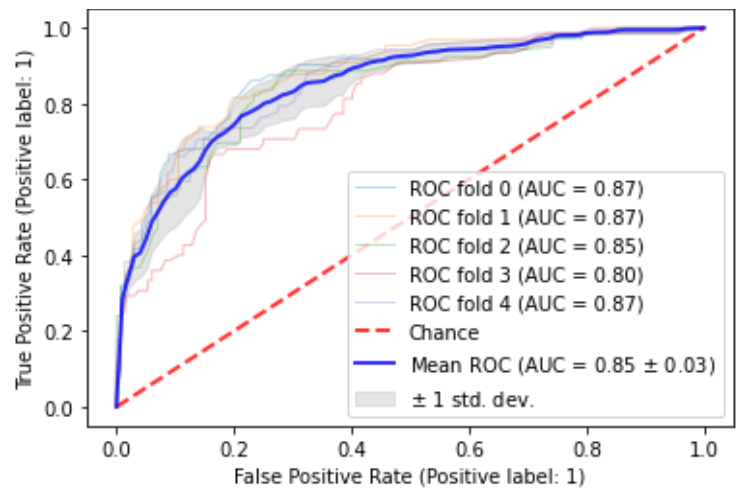

Figure 4. ROC curve for Depresjon 5-fold CV

## Conclusion

In this work, we presented a hybrid approach for development and deployment of a transfer learning-based depression screening tool. The primary issue with mental health specific disorders and their detection using AI is availability of limited data samples. Also, the distribution for the data collected using different wearables may differ. To mitigate both these challenges, we proposed a transfer learning method which uses the secondary dataset for training and a primary dataset for real time testing and deployment. The data modalities for both secondary and primary datasets is specifically maintained constant. The results demonstrate that such an approach can prove useful for developing digital healthcare solutions, typically in the applications where data scarcity is common.